\documentclass[11pt,a4paper]{article}
\usepackage[hyperref]{emnlp2020}

\usepackage[utf8]{inputenc} 
\usepackage[T1]{fontenc}    
\usepackage{microtype}
\usepackage{graphicx}
\usepackage{subfigure}
\usepackage{booktabs} 
\usepackage{comment}
\usepackage{mathtools}
\usepackage{multirow}
\usepackage{color}
\usepackage{float}
\usepackage{graphicx}
\usepackage{CJKutf8}
\usepackage{url}
\usepackage{bm}
\usepackage{parskip}
\usepackage{booktabs}       
\usepackage{amsmath, amsfonts, amsthm, amssymb}

\usepackage{amsmath, amsfonts}

\usepackage{times}
\usepackage{latexsym}

\aclfinalcopy 

\title{Improving  Robustness and  Generality of NLP Models \\ Using Disentangled  Representations}

\author{Jiawei Wu$^{\clubsuit}$, Xiaoya Li$^{\clubsuit}$,  Xiang Ao$^\blacklozenge$,  Yuxian Meng$^{\clubsuit}$, Fei Wu$^\spadesuit$ and Jiwei Li$^{\spadesuit \clubsuit}$ \\
$^\spadesuit$ Department of Computer Science and Technology, Zhejiang University\\
$^\blacklozenge$ Key Lab of Intelligent Information Processing of Chinese Academy of Sciences  \\
$^\clubsuit$ ShannonAI \\
 {\{jiawei\_wu, xiaoya\_li, yuxian\_meng, jiwei\_li\}}@shannonai.com\\
 aoxiang@ict.ac.cn, wufei@zju.edu.cn 
}

\date{}

\begin{document}
\maketitle

\begin{abstract}
Supervised 
neural networks, which first map an input $x$ to a single representation $z$, and then 
map $z$  
 to the output label $y$, 
have achieved remarkable success in a wide range of natural language processing (NLP) tasks. 
Despite their success, 
neural models lack for both robustness and generality: 
small perturbations to inputs can result in absolutely different outputs; 
the performance of a model trained on one domain drops drastically when tested on another domain. 

In this paper, we present  methods to improve robustness 
and generality of NLP models from the standpoint of disentangled representation learning. 
Instead of mapping $x$ to a single representation $z$, 
the proposed strategy maps $x$ to a set of representations $\{z_1,z_2,...,z_K\}$ while forcing them to be disentangled.
These representations are then mapped to different logits $l$s, the ensemble of which is used to make the final prediction $y$. 
We propose different methods to incorporate this idea into currently widely-used models, including 
adding an $L$2 regularizer on $z$s or adding Total Correlation (TC) under the framework of variational information bottleneck (VIB). 
We show that models trained with the proposed criteria provide better robustness 
and domain adaptation ability 
in a wide range of supervised learning tasks. 
\end{abstract}

\section{Introduction} 
Supervised neural networks 
have achieved remarkable success in a wide range of NLP tasks, such as language modeling \citep{xie2017data, bert, yinhan2019roberta, danqi2020spanbert, yuxian2019glyce}, machine reading comprehension \citep{bidaf, yu2018qanet}, and machine translation \citep{sutskever2014sequence,attentionisallyouneed, yuxian2019largenmt}. 
Despite  the success, 
neural models lack for both robustness and  generality and are extremely fragile:
the output label can be changed with a minor change of a single pixel \cite{szegedy2013intriguing,goodfellow2014explaining,nguyen2015deep,papernot2017practical,yuan2019adversarial} in an image
or a token in a document \cite{li2016understanding,papernot2016crafting,jia2017adversarial,zhao2017generating,ebrahimi2017hotflip,jia2019certified};
The model lacks for domain adaptation abilities \cite{mou2016transferable,daume2009frustratingly}: 
a model trained on one domain can hardly generalize to new test distributions \cite{fisch2019mrqa,levy2017zero}. 
Despite that
different avenues  have been proposed to address model robustness such as 
augmenting the training data using  rule-based lexical substitutions \cite{liang2017deep,ribeiro2018semantically} or paraphrase models \cite{iyyer2018adversarial}, 
building robust and domain-adaptive neural models remains  a challenge. 

In a standard supervised learning setup, a neural network model 
first
maps an input $x$ to a single vector $z=f(x)$.  
$z$ can be viewed as the hidden feature to represent $x$, and is transformed to its logit $l$ followed by a softmax operator to output the target label $y$. 
At training time, parameters involved in 
 mapping from $x\in X$ to $z$ then to $y$ are learned. At test time, the pretrained model makes a prediction
  when presented with a new instance $x'\in X'$. 
This methodology works well if $X$ and $X'$ come from exactly the same distribution, but significantly suffers if not.
This is because
the 
 implicit representation learned through
supervised  signals can easily and overfit to the training domain $X$, and
 the mapping function $f(x)$, which is trained only based on $X$, can be confused with out-of-domain features in  $x'$,
such as a lexical, pragmatic, and syntactic variation
not seen in the training set \cite{ettinger2017towards}.
We can also interpret the weakness of this  methodology 
from a domain adaptation point of view \cite{daume2006domain,daume2009frustratingly,tan2009adapting,patel2014visual}: 
it is crucial to separate  source-specific features,   
target-specific features and general features (features shared by sources and targets).
One of the most  naive strategies for domain adaptation is to ask the model to only use general features for test. 
In the standard $x\rightarrow z \rightarrow y$ setup, all features, including  source-specific,   
target-specific  and general features, are entangled in $z$. Due to the lack of interpretability \cite{li2015visualizing,linzen2016assessing,lei2016rationalizing,koh2017understanding} of neural models, it is impossible to disentangle them.

Inspired by recent work in disentangled representation learning \cite{bengio2013representation,kim2018factorvae,hjelm2018mutualvae,kumar2018variational,pmlr-v97-locatello19a},
we propose to improve  robustness 
and generality of NLP models
using disentangled  representations. 
Different from  mapping $x$ to a single representation $z$ and then to $y$, 
 the proposed strategy first maps $x$ to a set of distinct representations 
$Z=\{z_1,\cdots,z_K\}$, which are then individually projected to logits $l_1,\cdots,l_K$.
$l$s are ensembled to make the final prediction of $y$. 
In this setup,
we wish to make $z$s or $l$s to be disentangled from each other as much as possible, 
which potentially improves both robustness and  generality:
For the former, 
the decision  of $y$ is more immune to small
 changes in $x$
since  even though  small changes lead to significant changes in some $z$s or $l$s,
  others may remain invariant.
The ultimate 
influence on $y$ can be further regulated when $l$s are combined.
For the latter,  different $l$s have the potential to disentangle or partially disentangle
source-specific,   
target-specific  and general features. 

Practically, we propose two ways to disentangle representations: adding 
 an $L$2 regularizer 
or adding Total Correlation (TC) \citep{cover2012elements, ver2015maximally,steeg2017unsupervised,gao2018auto,chen2018isolating} under the framework of variational information bottleneck (VIB).
We show that models trained with the proposed criteria provide better robustness 
and domain adaptation ability 
 in a wide range of NLP tasks, with tiny or non-significant sacrifice on task-specific accuracies. 

In summary, the contributions of this paper are:
\begin{itemize}
    \item We present two methods to improve the robustness and generality of NLP models in the view of disentangled representation learning and the information bottleneck theory.
    \item Extensive experiments on domain adaptation and defense against adversarial attacks show that the proposed methods are able to provide better robustness compared with conventional task-specific models, which indicates the effectiveness of the theory of information bottleneck and disentangled representation learning for NLP tasks.
\end{itemize}

The rest of this paper is organized as follows: we present related work in Section 2. Models are detailed in Section 3 and Section 4. We present experimental results and analysis  in Section 5, followed by a brief conclusion in Section 6. 
\section{Related Work}
\subsection{Learning Disentangled Representations} 
Disentangled representation learning was first proposed by \citet{bengio2013representation}.
InfoGan \citep{chen2016infogan} disentangled the representation by maximizing the mutual information between a small subset of the GAN's noise latent variables and the observation.
\citet{kim2018factorvae} learned disentangled representations in VAE, by encouraging the distribution of representations to be factorial and hence independent across the dimensions.
\citet{hjelm2018mutualvae}  
learned disentangled representations by simultaneously estimating and maximizing the mutual information between input data and learned high-level representations.
\citet{chen2018isolating} proposed $\beta$-TCVAE, encouraging the model to find statistically independent factors in the data distribution by imposing a {\it total correlation} (TC) penalty. Similarly, \citet{kumar2018variational} learned disentangled latents from unlabeled observations by introducing a regularizer over the induced prior.

\subsection{The Information Bottleneck  Principle}
The Information Bottleneck (IB) principle was first proposed by \citet{info2000bottleneck}. 
It treats the supervised learning task as an optimization problem that squeezes the information from an input about the output through an information bottleneck. 
In information bottleneck, 
 the mutual information $I(X; Y)$ is used  as the measurement  of the relevant information  between  $x$ and the output $y$. 
\citet{tishby2015dl,shwartz2017opening} proposed to use it as a theoretical tool for analyzing and understanding representations  in deep neural networks. 
 \citet{alemi2016vib} proposed a deep variational version of the IB principle (VIB) to allow for using deep neural networks to parameterize the distributions.
In the field  of NLP,  not much attention has been attached to the Information Bottleneck principle. 
\citet{li2019specializing} 
 proposed to extract specific information
 for different tasks (which are defined in the output $y$)
  from pretrained word embeddings using VIB. 
Less relevant work is from \citet{kong2019mutual}, 
which proposed a  
 self-supervised objective that maximizes the mutual information between 
global sentence representations and $n$-grams in the sentence.

\subsection{Domain Adaptation in NLP}
Domain adaptation evalutes the model's ability of generalization across domains, for which many efforts have been devoted to designing more powerful cross-domain models \citep{daume2009frustratingly,kim-etal-2015-new,LEE18.878,adel2017unsupervised,yang-etal-2018-design,Ruder2019Neural}. 
\citet{AAAI1612443} proposed CORAL, a method that minimizes domain shift by aligning the second-order statistics of source and target distributions without even requiring any target labels;
\citet{lin-lu-2018-neural} added domain-adaptive layers on top of the model; 
\citet{jia2019cross} used cross-domain language models as a bridge cross-domains for domain adaptation. 
\citet{li-etal-2019-transferable,du2020adversarial} applied adversarial learning to learn cross-domain models for the task of sentiment analysis.
For machine translation, the core idea is to utilize large available parallel data for training NMT models and adapt them to  domains with small data \citep{chu2017empirical}, where data augmentation \citep{sennrich2016back-translation,ul-haq-etal-2020-improving}, meta-learning \citep{gu2018meta} and finetuning methods \citep{Luong-Manning:iwslt15,freitag2016fast,Dakwale2017FineTuningFN} are proposed to achieve this goal.

\subsection{Defense against Adversarial Attacks in NLP}
Deep neural networks are fragile when attacked by adversarial examples \citep{NIPS2014_5423,pmlr-v70-arjovsky17a,mirza2014conditional}. In the context of NLP, 
\citet{ijcai2018-601} built a candidate pool that includes adversarial examples, and used the method of Fast Gradient Sign Method (FGSM) \citep{goodfellow2014explaining} to select a candidate word for replacement. \citet{7795300} showed that the forward derivative \citep{7467366} can be used to produce adversarial sequences manipulating both the sequence output and  classification predictions made by an RNN. \citet{liang2017deep} designed three perturbation strategies for word-level attack --- insertion, modification and removal. 
\citet{miyato2016adversarial,ijcai2018-601,Zhu2020FreeLB,zhou2020defense} restricted the directions of perturbations toward the existing words in the input embedding space. 
 \citet{ebrahimi2017hotflip} proposed a novel token transformation method by computing derivatives with respect to a few character-edit operations. Other methods either generate certified defenses \citep{jia2019certified,huang2019achieving,Shi2020Robustness}, or generate examples that maintain lexical correctness, grammatical correctness and semantic similarity \citep{ren-etal-2019-generating}.

\section{Adding $L_2$ Regularizer on $Z$}
Here, we  present our first attempt to learn disentangled representations with an $L_2$  regularizer. 
We first map the input $x$ 
 to multiple  representations $Z = \{z_1,z_2,...,z_K\}$ and we wish different $z$s to be disentangled. 
To obtain $Z$, 
  we can use independent sets of parameters of 
RNNs \cite{hochreiter1997long,mikolov2010recurrent}, CNNs \cite{krizhevsky2012imagenet,kalchbrenner2014convolutional} or Transformers \cite{attentionisallyouneed}.
This actually mimics the idea  of the model ensemble. 
To avoid the parameter and memory intensity in the ensemble setup, we adopt the following simple method: 
we first map $x$ to a single vector representation $z$ using RNNs or CNNs. Next, we separate sub-representations from $z$ using distinct projection matrices, 
each of which tries to capture a certain aspect of features, given as follows:
\begin{equation}
  \begin{aligned}
    z_i=W_i z,\; i=1,\cdots,K
  \end{aligned}
\end{equation}
where $z$, $z_i\in \mathbb{R}^{d\times 1}$, 
 $W_i\in \mathbb{R}^{d\times d}$, 
and 
 $K$ is the number of  disentangled representations. 
 
 To make sure that these sub-representations actually disentangle, we enforce a regularizer on the  $L_2$ distance between each pair of them:
\begin{equation}
    \mathcal{L}_\text{reg}=\sum_{ij}\|z_i-z_j\|^2
\end{equation} 
The regularizer assumes that the distance between representations in the Euclidean space is in accordance with the distinctiveness between features that are the most salient for predictions. Each $z_i$ is next mapped to a logit $l_i$ as follows:
\begin{equation}
l_i = W \cdot z_i,\; i=1,\cdots,K
\end{equation}
where $W\in \mathbb{R}^{T\times d}$ and $T$ denotes the number of predefined classes for the supervised learning task. 
Next we aggregate the weighted logits into a single final logit $l =\sum \alpha_i l_i$, where $\alpha_i$ is the weight associated with $l_i$. 
$\alpha_i$ can be computed using the softmax operator by introducing a learnable parameter $w_a\in\mathbb{R}^{d\times 1}$: 
\begin{equation}
  \bm{\alpha} = \texttt{softmax}([z_1^\top w_a,\cdots,z_K^\top w_a])
\end{equation}
Combining the  cross entropy loss with golden label  $\hat{y}$ and the $L_2$ regularizer on $Z$, we can obtain the final training objective as follow:
\begin{equation}
  \mathcal{L}_\text{separare}=\text{CE}(\texttt{softmax}(l),\hat{y})+\beta\mathcal{L}_\text{reg}
\end{equation}
$\beta$ is the hyper-parameter controlling the weight of the regularizer. 
The method  can be adapted to any neural network. Albeit simple, this model has significantly better ability of 
learning disentangled features
and and is less prone to adversarial attacks, as we will show in the experiments later.

\section{Variational Information Bottleneck with Total Correlation}
Many
recent works \citep{alemi2016vib,Higgins2017betaVAELB,burgess2018understanding} 
 have shown that the information bottleneck is
  more suitable for learning  robust and general features than task-specific end-to-end models, due to the flexibility 
 provided by its learned structure. 
Here we first go through the preliminaries of the variational information bottleneck (VIB) \cite{alemi2016vib}, and then detail how it can be adapted for learning disentangled representations by 
adding a 
 Total Correlation (TC) regularizer \citep{ver2015maximally,steeg2017unsupervised,gao2018auto}.
 
\subsection{Variational Information Bottleneck}
Let $p(z|x)$ denote an encoding of $x$, which maps $x$ to representations $z$. 
The key point of IB is to learn an encoding that is maximally informative about our target
$Y$, measured by the mutual information between $z$ and the target $y$, denoted by $I(y,z)$. 
Unfortunately, only modeling  $I(y,z)$ is not enough since the model can always make $z=x$ to  ensure the maximally informative representation, which is not helpful for learning general features.
Instead, we need to find the best $z$  subject to a constraint on
its complexity, leading to the penalty on  the mutual information between $x$ and $z$. 
The objective for IB is thus given as follows: 
\begin{equation}
    \mathcal{L}_\text{IB}=I(z,y;\bm{\theta})-\beta I(z,x;\bm{\theta})
    \label{eq1}
\end{equation}
where $\beta$ controls the trade-off between $I(z,y)$ and $I(z,x)$. 
 Intuitively, the first term encourages $z$ to be predictive of $y$ and the second term enforces $z$ to be concisely representative of $x$.\footnote{It is worth noting that Eq.\ref{eq1} resembles the form of $\beta$-VAE \citep{Higgins2017betaVAELB}, an unsupervised model for learning disentangled representations modified upon the Variational Autoencoder (VAE) \citep{kingma2013auto}. 
\citet{burgess2018understanding} showed from an information bottleneck view that $\beta$-VAE mimics the behavior of information bottleneck and learns to disentangle representations.}

By leaving details to the appendix,
we can obtain the lower bound of $I(z,y)$  and the upper bound of $I(z,x)$: 
\begin{equation}
  \begin{aligned}
    &I(z,y)\ge \int p(x)p(y|x)p(z|x)\log q(y|z)\ \mathrm{d}x\mathrm{d}y\mathrm{d}z \\
    &    I(z,x)\le\int  p(x)p(z|x)\log \frac{p(z|x)}{r(z)} \ \mathrm{d}x\mathrm{d}z 
  \end{aligned}
\end{equation}
where $q(y|z)$ and $r(z)$ are variational approximations to $p(y|z)$ and $p(z)$ respectively. We can immediately have the lower bound of Eq.\ref{eq1}:
\begin{equation}
  \begin{aligned}
    &I(Z,Y)-\beta I(Z,X)\ge \\
    & \int p(x)p(y|x)p(z|x)\log q(y|z)\ \mathrm{d}x\mathrm{d}y\mathrm{d}z \\
    &-\beta \int  p(x)p(z|x)\log \frac{p(z|x)}{r(z)}\ \mathrm{d}x\mathrm{d}z=\mathcal{L}_\text{VIB}
  \end{aligned}
\end{equation}
In order to compute this in practice, we approximate $p(x,y)$ using  the empirical data distribution $p(x,y)=\frac{1}{N}\sum_{n=1}^N\delta_{x_n}(x)\delta_{y_n}(y)$, leading to:
\begin{equation}
  \begin{aligned}
    \mathcal{L}_\text{VIB}&\approx \frac{1}{N}\sum_{n=1}^N\left[\int p(z|x_n)\log q(y_n|z)\ \mathrm{d}z-\right.\\
    &\left. \beta p(z|x_n)\log\frac{p(z|x_n)}{r(z)}\right]
  \end{aligned}
\end{equation}
By using the reparameterization trick \citep{kingma2013auto} to rewrite $p(z|x) \mathrm{d}z=p(\epsilon)\ \mathrm{d}\epsilon$, where $z=f(x,\epsilon)$ is a deterministic function of $x$ and the Guassian random variable $\epsilon$, we put everything together to the following objective:
\begin{equation}
\begin{aligned}
  \mathcal{L}_\text{VIB}=&\frac{1}{N}\sum_{n=1}^N\mathbb{E}_{p(\mathbf{\epsilon})}[-\log q(y_n|f(x_n,\mathbf{\epsilon}))]+\\&\beta \mathcal{D}_\text{KL}(p(z|x_n),r(z))
  \label{eq3}
\end{aligned}
\end{equation} $p(z|x)$ is set to $\mathcal{N}(z|f^\mu_e(x),f^\Sigma_e(x))$ where $f_e$ is an MLP of mapping the input $x$ to a stochastic encoding $z$. The output dimension of $f_e$ is $2D$, where the first $D$ outputs encode $\mu$ and the remaining $D$ outputs encode $\sigma$. Then we sample $\epsilon\sim \mathcal{N}(\mathbf{0},\mathbf{I})$ and combine them together $z=\mu+\epsilon\cdot \sigma$.
We treat $r(z)=\mathcal{N}(z|\mathbf{0},\mathbf{I})$ and $q(y|z)$ as a softmax classifier.
Eq. \ref{eq3} can be trained by directly backpropagating through examples and the gradient is an unbiased estimate of the true gradient.

\subsection{VIB+TC: VIB with Total Correlation}
While VIB provides a neat way of parameterizing the information bottleneck approach and efficiently training the model with the reparameterization trick,  the learned  representations only contain the minimal statistics required to predict the target label, it does not 
immediately 
have the ability to {disentangle} the learned representations.
To tackle this issue,  another regularizer is added,  the Total Correlation (TC) \citep{ver2015maximally,steeg2017unsupervised,gao2018auto}, to disentangle $z$: 
\begin{equation}
  \begin{aligned}
      &\text{TC}(z_1, .., z_K|x) = \sum_{i=1}^{K} H(z_i|x) - H(z_1,\cdots,z_K|x) \\
      &= \mathcal{D}_\text{KL} \left[p(z_1,\cdots,z_K|x), \prod_{i=1}^{K} p(z_i|x)   \right]
  \end{aligned}
  \end{equation}
The TC term  measures the dependence between $p(z_i|x)$s. The penalty on TC forces the model to find statistically independent factors in the features. 
In particular, $\text{TC}(z_1, .., z_K|x)$ is zero if and only if all $p(z_i|x)$s are independent, in which case we say that they are disentangled. 
Thus, the training objective is defined as follows: 
\begin{equation}
\begin{aligned}
     &\mathcal{L}_\text{VIB+TC}
      = \frac{1}{N} \sum_{i=1}^{K} \sum_{n=1}^{N} \mathbb{E}_{p(\epsilon)} [ {- \log q(y_n|z_i)}\\
      &{+ \beta \mathcal{D}_\text{KL} ( p(z_i|x_n), r(z_i))} ] \\
     & + \lambda \mathcal{D}_\text{KL}\left[p(z_1, ..., z_K|x), \prod_{i=1}^{K} p(z_i|x)  \right]
     \label{eq6}
\end{aligned}
\end{equation}
where $\beta$ and $\lambda$ are  a hyper-parameters to adjust the trade-off between these two factors. 
$p(z_i|x)$ is set to $\mathcal{N}(z|f^\mu_{e,i}(x),f^\Sigma_{e,i}(x))$, in a similar way to $p(z|x)$ except that 
$f^\mu_{e,i}(x)$ and $f^\Sigma_{e,i}$ are scalars. 
Eq.\ref{eq6} can also be directly trained with an unbiased estimate of the true gradient. 

\section{Experiments}
In this section, we describe experimental results. We conduct experiments in two NLP subfields: domain adaptation and defense against adversarial attacks. 
\subsection{Domain Adaptation}
The goal of
domain adaptation tasks is to test whether a model trained in one domain (source-domain) can work well when test in another domain (target-domain).
In the domain adaptation setup, there should be at least labeled source-domain  data for training and labeled target-domain  data for test.
Setups can be different regarding whether there is also a small amount of   labeled target-domain  data for training  or 
unlabeled target-domain data for unsupervised training \cite{jia2019cross}. 
In this paper, we adopt the most naive setting where there is neither labeled nor unlabeled
 target-domain  data for training to straightforwardly test a model's ability for domain adaptation. 
We perform experiments on the following 
domain adaptation tasks: named entity recognition (NER), part-of-speech tagging (POS), machine translation (MT) and text classification (CLS). 
 The $L_2$ regularizer, VIB and VIB+TC models
  are built on top of representations of the last layer for fair comparison.

\begin{table*}[t]
  \centering
  \begin{tabular}{cccccc}
    \toprule
    {\bf Method} & NER & POS & MT & CLS-sentiment & CLS-deception\\
    \midrule
    Baseline & 97.88 & 90.12 & 34.61 & 87.4 & 87.5\\
    VIB & 98.02$_{+0.14}$ & 90.85$_{+0.73}$ & 34.90$_{+0.29}$ & 88.5$_{+1.1}$ & 88.6$_{+1.1}$\\
    VIB+TC & {\bf 98.33}$_{+0.45}$ & {\bf 91.43}$_{+1.31}$ & {\bf 35.31}$_{+0.70}$ & {\bf 89.8}$_{+2.4}$ & {\bf 89.3}$_{+1.8}$\\
    Regularizer & 98.21$_{+0.33}$ & 91.30$_{+1.18}$  & 35.13$_{+0.52}$ & 89.2$_{+1.8}$  & 88.7$_{+1.2}$\\
    \bottomrule
  \end{tabular}
  \caption{Results for domain adaptation. The evaluation metric for NER, POS and CLS is accuracy, and that for MT is the BLEU score \citep{papineni-etal-2002-bleu}.}
  \label{table1}
\end{table*}

\paragraph{NER}
For the task of NER, we followed
the setup 
 in \newcite{daume2009frustratingly} and used 
 the
  ACE06 dataset as the source domain and the
CoNLL 2003 NER data as the target domain. 
The training dataset of 
  ACE06 contains 256,145  examples, and the dev and test datasets from CoNLL03 respectively contains  5,258 and 8,806 examples. 
For evaluation, we followed \newcite{daume2009frustratingly} and report only on label accuracy. 
We used the MRC-NER model as the baseline \cite{li2019unified},  which achieves SOTA performances on a wide range of NER tasks.\footnote{MRC-NER transforms tagging tasks to MRC-style span prediction tasks, which 
first
concatenates category descriptions with texts to tag. The concatenation is then fed to the BERT-large model \cite{bert} to predict the corresponding start index and end index of the entity. }
All models are trained using using Adam \citep{kingma2014adam} with $\beta=(0.9,0.98)$, $\epsilon=10^{-6}$, a polynomial learning rate schedule, warmup up for 4K steps and weight decay with $10^{-3}$.\footnote{We optimize the learning rate in the range {1e-5, 2e-5, 3e-5, 5e-5} with dropout rate set to 0.2.}

\paragraph{POS}
For the task of POS, we 
followed the setup 
 in \newcite{daume2009frustratingly}.
 The source domain is 
  the WSJ portion of
the Penn Treebank, containing 950,028 training examples.
The target domain is
PubMed, with the dev and test sets respectively containing  1,987 and 14,554  examples. 
We used the BERT-large model as the backbone. The model is optimized using   Adam  \citep{kingma2014adam}. 

\paragraph{Machine Translation}
We used the WMT 2014  English-German dataset for training, which contains about 4.5 million sentence pairs. 
We used the Tedtalk dataset \cite{duh2018multitarget} for test. 
We use the Transformer-base model \cite{vaswani2017transformer} as the backbone, where the encoder and decoder respectively have 6 layers. 
Sentences are encoded using BPE \citep{sennrich-etal-2016-neural}, which has a shared source target vocabulary of about 37000 tokens.
For fair comparison, we used the Adam optimizer \citep{kingma2014adam} with $\beta_1$ = 0.9, $\beta_2$ = 0.98 
and $\epsilon$ = $10^{-9}$ for all models. 
For the base setup, following \newcite{vaswani2017transformer},
the dimensionality of inputs and outputs  $d_\text{model}$ is set to 512, and the inner-layer has dimensionality $d_\text{ff}$ is set to 2,048.

\paragraph{Text Classification}
For text classification, we used two datasets.
The first dataset we consider is the sentiment analysis on reviews. We used the 450K Yelp reviews for training and $\sim$ 3k Amazon reviews for test \cite{li2018delete}.
The task is transformed to a binary classification task to decide whether a review is of positive or negative sentiment. 
 We also used the deceptive opinion spam detection dataset \cite{li2014towards}, a binary text classification task to classify whether a review is fake or not.
We used the  hotel reviews for training, which consists of 800 reviews in total from customers, and used the 400 restaurant reviews for test. 
For baselines, we used the BERT-large model \citep{devlin2018bert} as the backbone, where the \texttt{[cls]} is first mapped to a scalar and then output to a sigmoid function. 
We report accuracy on the test set. 

\paragraph{Results} Results for domain adaptation are shown in Table \ref{table1}. As can be seen, for all tasks, VIB+TC performs best among all four models, followed by the proposed $L2$ regularizer model,  next followed by the VIB model without disentanglement. 
The vanilla VIB model outperforms the baseline supervised model. This is because 
the VIB model maps an input to multiple representations, and this operation  to some degree separates features in a natural way. 
The $L2$ regularizer method consistently outperforms VIB and underperforms VIB+TC.
This is because VIB+TC
 uses the TC term to  disentangle features deliberately, and the vanilla VIB model does not have this property. 
 Experimental results demonstrate the importance of learning disentangled features in domain adaptation.  
 
\begin{table*}[t]
  \centering
  \footnotesize
  \resizebox{\textwidth}{26mm}{
  \begin{tabular}{ccccccccccccc}
    \toprule
    \multicolumn{13}{c}{{\it IMDB}}\\
    \multirow{2}{*}{\bf Method} & \multicolumn{4}{c}{{BoW}} & \multicolumn{4}{c}{{CNN}} & \multicolumn{4}{c}{{LSTM}}\\
    &  Clean & PWWS& GA$_\text{w/LM}$& GA$_\text{w/o LM}$ & Clean & PWWS& GA$_\text{w/LM}$& GA$_\text{w/o LM}$ & Clean & PWWS& GA$_\text{w/LM}$& GA$_\text{w/o LM}$   \\
    \midrule
    Orig. & 88.7 & 12.4 & 2.1 & 0.7 & 90.0 & 18.1 & 4.2 & 2.0 & 89.7 & 1.4 & 2.5 & 0.1 \\
    VIB &   88.6 &	22.4&	19.0	&11.5	&	89.3&	36.1&	34.7&	13.1&		88.9&	14.2&	31.4&	7.6\\
    VIB+TC & 89.1&	{\bf 26.5}	&{\bf 21.4}	&{\bf 19.5}	&	89.5&	{\bf 40.2}&	{\bf 39.0}&	18.6	&	89.6&	16.9	&{\bf 33.0}	&{\bf  10.4}\\
    Regularizer & {\bf 90.1}&	17.1&	7.2	&3.7	&	{\bf 90.6}&	21.4&	15.1&	8.9	&	{\bf 90.1}&	3.1	&15.3	&5.8\\
    \hline
    \hline 
    \multicolumn{13}{c}{{\it AGNews}}\\
    \multirow{2}{*}{\bf Method} & \multicolumn{4}{c}{{BoW}} & \multicolumn{4}{c}{{CNN}} & \multicolumn{4}{c}{{LSTM}}\\
    &  Clean & PWWS& GA$_\text{w/LM}$& GA$_\text{w/o LM}$ & Clean & PWWS& GA$_\text{w/LM}$& GA$_\text{w/o LM}$ & Clean & PWWS& GA$_\text{w/LM}$& GA$_\text{w/o LM}$   \\
    \midrule
    Orig. &{\bf 88.4}&	45.2&	58.3	&19.5	&	89.2&	37.8	&45.7&	12.5&		92.4&	46.8&	48.7&	9.4    \\
    VIB &   87.9&	57.4&	64.5&	32.2&		88.5&	50.4&	54.7	&21.0&		91.4	&57.6	&59.7	&19.2    \\
    VIB+TC &  87.6&	{\bf 61.4}	&{\bf 72.1}&	{\bf 34.5}&		89.0&	{\bf 54.3}&	{\bf 59.2}&	{\bf 25.4}&		92.5&	{\bf 61.1}&	{\bf 65.4}	&{\bf 21.1}    \\
    Regularizer & {\bf 88.4}&	50.1&	65.0&	25.4		&{\bf 89.4}&	43.5&	50.2&	17.5	&	{\bf 92.8}	&50.3&	52.1&	11.0    \\
    \bottomrule
  \end{tabular}}
  \caption{Results for the IMDB and AGNews datasets. \texttt{Orig.} stands for the original baseline, on which all other methods are based. Accuracy is reported for comparison.}
  \label{table2}
\end{table*}

\begin{table}[t]
  \centering
  \small
  \begin{tabular}{ccccc}
    \toprule
    \multicolumn{5}{c}{{\it SNLI}}\\
    {\bf Method} & Clean & PWWS & GA$_\text{w/LM}$& GA$_\text{w/o LM}$\\
    \midrule
    Orig. & {\bf 90.5}&	43.1&	55.6	&21.4    \\
    VIB & 89.4&	56.5&	62.4&	35.9    \\
    VIB+TC & 89.9	&{\bf 62.4}	&{\bf 67.0}&	{\bf 41.3}    \\
    Regularizer & 90.4&	48.1&	59.6	&27.2    \\
    \bottomrule
  \end{tabular}
  \caption{Results for the SNLI dataset. We report accuracy for all models.}
  \label{table3}
\end{table}

\subsection{Defense Against Adversarial Attacks}
We evaluate the proposed methods on tasks for defense against adversarial attacks. 
We conduct experiments on the tasks of text classification and natural language inference in defense against two recently proposed attacks: PWWS and GA. 
PWWS \cite{ren2019generating}, short for Probability Weighted
Word Saliency, 
 performs 
text adversarial attacks based on word substitutions with synonyms. The word replacement order is determined by  both word saliency and prediction probability. 
GA \cite{alzantot2018generating} uses language models to 
remove candidate substitute words that do not fit within the context. 
We report the accuracy under GA attacks for both with
and without using the LM.

Following \newcite{zhou2020defense}, 
for text classification, we use two datasets,  IMDB (Internet Movie Database) and AG News corpus \citep{10.1145/1060745.1060764}. 
IMDB contains 50, 000 movie reviews for binary (positive v.s. negative)
sentiment classification, and AGNews contains roughly 30, 000 news articles for 4-class classification. 
We use three base models: bag-of-words models, 
CNNs and two-layer LSTMs.
The bag-of-words
model first averages the embeddings of constituent words of the input, and then passes the average embedding to a feedforward network to get a 100$d$ vector.
The vector is then mapped to the final logit. 
CNNs and LSTMs are used to map input text sequences to vectors, which are fed to \texttt{sigmoid} for IMDB and \texttt{softmax} for  AGNews. 

For natural language inference, we conduct experiments on the  Stanford Natural Language Inference (SNLI)  corpus \citep{snli-emnlp2015}. The dataset consists of 570, 000 English sentence pairs. The task is transformed to a 3-class classification problem, giving one of the entailment, contradiction, or 
neutral label to the sentence pair. 
All models use BERT as backbones and are trained on the \texttt{CrossEntropy} loss, and their hyper-parameters are tuned on the validation set.

\paragraph{Results} Table \ref{table2} shows  results for the IMDB and AGNews datasets, and Table \ref{table3} shows  results for the SNLI dataset.  
When tested on the clean dataset where no attack is performed,
variational methods, i.e., VIB and VIB+TC, underperform the baseline model. This is in line with our expectation:
because of the necessity of modeling the KL divergence between $z$ and $x$, 
 the variational methods do not gets to label prediction as straightly as supervised learning models. 
But  variational methods significantly outperform supervised baselines when attacks are performed, which is because of the flexibility offered by the disentangled latent representations. VIB+TC outperforms VIB due to the disentanglement introduced by TC when attacks are present. 
As expected, the $L2$ regularizer model outperforms the baseline model in terms of robustness in defense against adversarial attacks. 
It is also interesting that with $L2$ regularizer, the model
performs at least comparable to, and sometimes outperforms the baseline in the setup without adversarial attacks, which demonstrates that disentangled representations can also help alleviate overfitting, leading to better performances.

\subsection{Ablation Studies} 

\begin{table}[t]
  \centering
  \begin{tabular}{ccccc}
    \toprule
    \multicolumn{5}{c}{{\it IMDB}}\\
    {\bf $\beta$} & Clean & PWWS & GA$_\text{w/LM}$& GA$_\text{w/o LM}$\\
    \midrule
    0 & 90.0&	18.1&	4.2&2.0    \\
    0.05 & 90.3 &19.7	&11.4 &6.2   \\
    0.10 & {\bf 90.6}	&21.4	&15.1 &	{\bf 8.9}    \\
    0.15 & 90.1&	{\bf 22.7}&{\bf 17.2}		& 8.8   \\
    0.20 &89.5 &22.3	&	15.5&7.5    \\
    0.25 &88.7 &20.6	&13.2		&6.5    \\
    0.30 & 87.9&18.9	&	11.6	&4.5    \\
    \bottomrule
  \end{tabular}
  \caption{Results of varying the hyperparamter $\beta$ in the Regularizer method. Accuracy is reported for comparison. The backbone model is CNN.}
  \label{table4}
\end{table}

\begin{table}[t]
  \centering
  \begin{tabular}{ccccc}
    \toprule
    \multicolumn{5}{c}{{\it IMDB}}\\
    {\bf $\gamma$} & Clean & PWWS & GA$_\text{w/LM}$& GA$_\text{w/o LM}$\\
    \midrule
    0 & 89.3&36.1	&34.7	&13.1    \\
    0.05 & {\bf 89.5} &37.9	& 37.2&15.1   \\
    0.10 &{\bf 89.5} &39.1	& {\bf 39.0}&17.6    \\
    0.15 &{\bf 89.5} &{\bf 40.2} &38.8	&{\bf 18.6}    \\
    0.20 &88.4 &38.5	&	38.0&17.9    \\
    0.25 & 87.4&37.2	&	36.8	&16.6    \\
    0.30 & 86.2&36.5	&	36.5&15.2    \\
    \bottomrule
  \end{tabular}
  \caption{Results of varying the hyperparamter $\gamma$ in the VIB+TC method. Accuracy is reported for comparison. The backbone model is CNN.}
  \label{table5}
\end{table}

\begin{figure*}[t]
  \centering
  \includegraphics[scale=0.32]{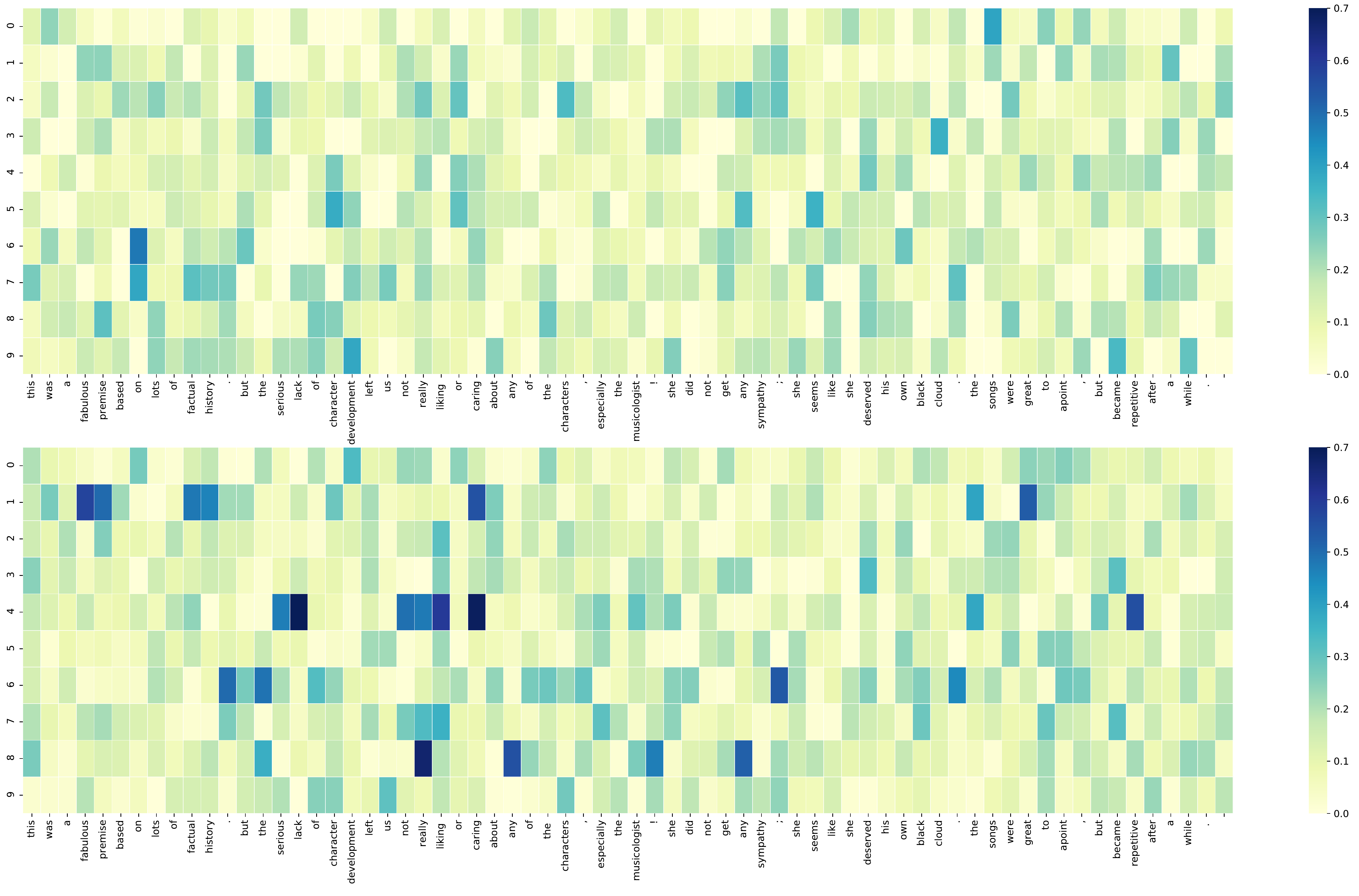}
  \caption{Heatmaps  for models without (top) and with (bottom) $L_2$ regularizer.}
  \label{fig1}
\end{figure*}

Next, we  explore how the strength of the regularization terms in VIB+TC and Regularizer affects  performances. Specifically, we vary the coefficient hyperparamter $\beta$ in Regularizer and the $\gamma$ in VIB+TC to show their influences on defending against adversarial attacks. We use the IMDB dataset for evaluation and use CNNs as baselines, and for each setting, we tune all other hyperparamters on the validation set. 

Results are shown in Table \ref{table4} and Table \ref{table5}.  As can be seen from the tables, when these two hyperparamters are around $0.1\sim 0.15$, the best results are achieved. For both methods, the performance first rises when increasing the hyperparameter value, and then drops as we continue increasing it. Besides, the difference between the best result and the worst result in the same model is surprisingly large (e.g., for the PWWS attack, the difference is 4.6 for Regularizer and 4.1 for VIB+TC), indicating the importance and the sensitivity of the introduced regularizers.

\subsection{Visualization}
It would be interesting to visualize how the disentangled $z$s encode the information of different parts of the input. 
Unlike feature-based models like SVMs, it's intrinsically hard to measure the influence of units 
of one layer on another layer in an neural architecture \citep{zeiler2014visualizing,yosinski2014transferable,bau2017network,koh2017understanding}. 
We turn to the first-derivative saliency method, a widely used  tool to visualize the influence of a change in the input on the model's predictions 
\citep{erhan2009visualizing,simonyan2013deep,li2015visualizing}. 
Specially, we want to visualize the influence of an input token $e$ on the $j$-th dimension of $z_i$, denoted by $z_i^j$. 
In the case of deep neural models, 
$z_i^j$ 
is a highly non-linear function of $e$. 
 The first-derivative saliency method approximates $z_i^j$ with a linear function of $e$ by computing the first-order Taylor expansion
\begin{equation}
z_i^j \approx w_i^j(e)^\top e+b
\end{equation}
where $w_i^j(e)$ is the derivative of $z_i^j$ with respect to the embedding $e$.
\begin{equation}
w_i^j(e)=\frac{\partial(z_i^j)}{\partial e}\Big|_e
\end{equation}
The magnitude (absolute value) of the derivative indicates 
the sensitiveness of  the final decision to the change in one particular word embedding, 
telling us how much one specific token contributes to $z$.
By summing over $j$, the influence of $e$ on $z_i$ is given as follows:
\begin{equation}
S_i(e)=\sum_j |w_i^j(e)|
\end{equation}
Figure 
 plots the heatmaps of $S_i(e)$ with respect to word input vectors
for 
models with and without the TC regularizer. 
As can be seen, by pushing representations to be disentangled, different representations are able to encode separate meanings of texts: $z_1$ tends to encode more positive information while $z_4$ tends to encode negative information.
This ability for feature separation and meaning clustering 
potentially improves the model's robustness.

\section{Conclusion}
In this paper, we present  methods to improve the robustness and generality on various NLP tasks in the perspective of the information bottleneck theory and disentangled representation learning. In particular, we find the two variational methods VIB and VIB+TC perform well on cross domain and adversarial attacks defense tasks. The proposed simple yet effective end-to-end method of learning disentangled representations 
with $L_2$ regularizer 
performs comparably well on cross-domain tasks, while better than vanilla non-disentangled models on adversarial attacks defense tasks, which shows the effectiveness of disentangled representations.

\bibliography{emnlp2020}

\begin{thebibliography}{107}
\expandafter\ifx\csname natexlab\endcsname\relax\def\natexlab#1{#1}\fi

\bibitem[{Adel et~al.(2017)Adel, Zhao, and Wong}]{adel2017unsupervised}
Tameem Adel, Han Zhao, and Alexander Wong. 2017.
\newblock Unsupervised domain adaptation with a relaxed covariate shift
  assumption.
\newblock In \emph{Thirty-First AAAI Conference on Artificial Intelligence}.

\bibitem[{Alemi et~al.(2016)Alemi, Fischer, Dillon, and Murphy}]{alemi2016vib}
Alexander~A Alemi, Ian Fischer, Joshua~V Dillon, and Kevin Murphy. 2016.
\newblock Deep variational information bottleneck.
\newblock \emph{arXiv preprint arXiv:1612.00410}.

\bibitem[{Alzantot et~al.(2018)Alzantot, Sharma, Elgohary, Ho, Srivastava, and
  Chang}]{alzantot2018generating}
Moustafa Alzantot, Yash Sharma, Ahmed Elgohary, Bo-Jhang Ho, Mani Srivastava,
  and Kai-Wei Chang. 2018.
\newblock Generating natural language adversarial examples.
\newblock \emph{arXiv preprint arXiv:1804.07998}.

\bibitem[{Arjovsky et~al.(2017)Arjovsky, Chintala, and
  Bottou}]{pmlr-v70-arjovsky17a}
Martin Arjovsky, Soumith Chintala, and L{\'e}on Bottou. 2017.
\newblock {W}asserstein generative adversarial networks.
\newblock volume~70 of \emph{Proceedings of Machine Learning Research}, pages
  214--223, International Convention Centre, Sydney, Australia. PMLR.

\bibitem[{Bau et~al.(2017)Bau, Zhou, Khosla, Oliva, and
  Torralba}]{bau2017network}
David Bau, Bolei Zhou, Aditya Khosla, Aude Oliva, and Antonio Torralba. 2017.
\newblock Network dissection: Quantifying interpretability of deep visual
  representations.
\newblock In \emph{Proceedings of the IEEE conference on computer vision and
  pattern recognition}, pages 6541--6549.

\bibitem[{Bengio et~al.(2013)Bengio, Courville, and
  Vincent}]{bengio2013representation}
Yoshua Bengio, Aaron Courville, and Pascal Vincent. 2013.
\newblock Representation learning: A review and new perspectives.
\newblock \emph{IEEE transactions on pattern analysis and machine
  intelligence}, 35(8):1798--1828.

\bibitem[{Bowman et~al.(2015)Bowman, Angeli, Potts, and
  Manning}]{snli-emnlp2015}
Samuel~R. Bowman, Gabor Angeli, Christopher Potts, and Christopher~D. Manning.
  2015.
\newblock A large annotated corpus for learning natural language inference.
\newblock In \emph{Proceedings of the 2015 Conference on Empirical Methods in
  Natural Language Processing (EMNLP)}. Association for Computational
  Linguistics.

\bibitem[{Burgess et~al.(2018)Burgess, Higgins, Pal, Matthey, Watters,
  Desjardins, and Lerchner}]{burgess2018understanding}
Christopher~P Burgess, Irina Higgins, Arka Pal, Loic Matthey, Nick Watters,
  Guillaume Desjardins, and Alexander Lerchner. 2018.
\newblock Understanding disentangling in $\beta$-vae.
\newblock \emph{arXiv preprint arXiv:1804.03599}.

\bibitem[{Chen et~al.(2018)Chen, Li, Grosse, and Duvenaud}]{chen2018isolating}
Ricky~TQ Chen, Xuechen Li, Roger~B Grosse, and David~K Duvenaud. 2018.
\newblock Isolating sources of disentanglement in variational autoencoders.
\newblock In \emph{Advances in Neural Information Processing Systems}, pages
  2610--2620.

\bibitem[{Chen et~al.(2016)Chen, Duan, Houthooft, Schulman, Sutskever, and
  Abbeel}]{chen2016infogan}
Xi~Chen, Yan Duan, Rein Houthooft, John Schulman, Ilya Sutskever, and Pieter
  Abbeel. 2016.
\newblock Infogan: Interpretable representation learning by information
  maximizing generative adversarial nets.
\newblock In \emph{Advances in neural information processing systems}, pages
  2172--2180.

\bibitem[{Chu et~al.(2017)Chu, Dabre, and Kurohashi}]{chu2017empirical}
Chenhui Chu, Raj Dabre, and Sadao Kurohashi. 2017.
\newblock An empirical comparison of domain adaptation methods for neural
  machine translation.
\newblock In \emph{Proceedings of the 55th Annual Meeting of the Association
  for Computational Linguistics (Volume 2: Short Papers)}, pages 385--391.

\bibitem[{Cover and Thomas(2012)}]{cover2012elements}
Thomas~M Cover and Joy~A Thomas. 2012.
\newblock \emph{Elements of information theory}.
\newblock John Wiley \& Sons.

\bibitem[{Dakwale(2017)}]{Dakwale2017FineTuningFN}
Praveen Dakwale. 2017.
\newblock Fine-tuning for neural machine translation with limited degradation
  across in-and out-of-domain data.

\bibitem[{Daum{\'e}~III(2009)}]{daume2009frustratingly}
Hal Daum{\'e}~III. 2009.
\newblock Frustratingly easy domain adaptation.
\newblock \emph{arXiv preprint arXiv:0907.1815}.

\bibitem[{Daume~III and Marcu(2006)}]{daume2006domain}
Hal Daume~III and Daniel Marcu. 2006.
\newblock Domain adaptation for statistical classifiers.
\newblock \emph{Journal of artificial Intelligence research}, 26:101--126.

\bibitem[{Del~Corso et~al.(2005)Del~Corso, Gull\'{\i}, and
  Romani}]{10.1145/1060745.1060764}
Gianna~M. Del~Corso, Antonio Gull\'{\i}, and Francesco Romani. 2005.
\newblock Ranking a stream of news.
\newblock WWW '05, page 97–106, New York, NY, USA. Association for Computing
  Machinery.

\bibitem[{Devlin et~al.(2018{\natexlab{a}})Devlin, Chang, Lee, and
  Toutanova}]{bert}
Jacob Devlin, Ming-Wei Chang, Kenton Lee, and Kristina Toutanova.
  2018{\natexlab{a}}.
\newblock Bert: Pre-training of deep bidirectional transformers for language
  understanding.
\newblock \emph{arXiv preprint arXiv:1810.04805}.

\bibitem[{Devlin et~al.(2018{\natexlab{b}})Devlin, Chang, Lee, and
  Toutanova}]{devlin2018bert}
Jacob Devlin, Ming-Wei Chang, Kenton Lee, and Kristina Toutanova.
  2018{\natexlab{b}}.
\newblock Bert: Pre-training of deep bidirectional transformers for language
  understanding.
\newblock \emph{arXiv preprint arXiv:1810.04805}.

\bibitem[{Du et~al.(2020)Du, Sun, Wang, Qi, and Liao}]{du2020adversarial}
Chunning Du, Haifeng Sun, Jingyu Wang, Qi~Qi, and Jianxin Liao. 2020.
\newblock Adversarial and domain-aware bert for cross-domain sentiment
  analysis.
\newblock In \emph{Proceedings of the 58th Annual Meeting of the Association
  for Computational Linguistics}, pages 4019--4028.

\bibitem[{Duh(2018)}]{duh2018multitarget}
Kevin Duh. 2018.
\newblock The multitarget ted talks task.

\bibitem[{Ebrahimi et~al.(2017)Ebrahimi, Rao, Lowd, and
  Dou}]{ebrahimi2017hotflip}
Javid Ebrahimi, Anyi Rao, Daniel Lowd, and Dejing Dou. 2017.
\newblock Hotflip: White-box adversarial examples for text classification.
\newblock \emph{arXiv preprint arXiv:1712.06751}.

\bibitem[{Erhan et~al.(2009)Erhan, Bengio, Courville, and
  Vincent}]{erhan2009visualizing}
Dumitru Erhan, Yoshua Bengio, Aaron Courville, and Pascal Vincent. 2009.
\newblock Visualizing higher-layer features of a deep network.
\newblock \emph{University of Montreal}, 1341(3):1.

\bibitem[{Ettinger et~al.(2017)Ettinger, Rao, Daum{\'e}~III, and
  Bender}]{ettinger2017towards}
Allyson Ettinger, Sudha Rao, Hal Daum{\'e}~III, and Emily~M Bender. 2017.
\newblock Towards linguistically generalizable nlp systems: A workshop and
  shared task.
\newblock \emph{arXiv preprint arXiv:1711.01505}.

\bibitem[{Fisch et~al.(2019)Fisch, Talmor, Jia, Seo, Choi, and
  Chen}]{fisch2019mrqa}
Adam Fisch, Alon Talmor, Robin Jia, Minjoon Seo, Eunsol Choi, and Danqi Chen.
  2019.
\newblock Mrqa 2019 shared task: Evaluating generalization in reading
  comprehension.
\newblock \emph{arXiv preprint arXiv:1910.09753}.

\bibitem[{Freitag and Al-Onaizan(2016)}]{freitag2016fast}
Markus Freitag and Yaser Al-Onaizan. 2016.
\newblock Fast domain adaptation for neural machine translation.
\newblock \emph{arXiv preprint arXiv:1612.06897}.

\bibitem[{Gao et~al.(2018)Gao, Brekelmans, Steeg, and Galstyan}]{gao2018auto}
Shuyang Gao, Rob Brekelmans, Greg~Ver Steeg, and Aram Galstyan. 2018.
\newblock Auto-encoding total correlation explanation.
\newblock \emph{arXiv preprint arXiv:1802.05822}.

\bibitem[{Goodfellow et~al.(2014{\natexlab{a}})Goodfellow, Pouget-Abadie,
  Mirza, Xu, Warde-Farley, Ozair, Courville, and Bengio}]{NIPS2014_5423}
Ian Goodfellow, Jean Pouget-Abadie, Mehdi Mirza, Bing Xu, David Warde-Farley,
  Sherjil Ozair, Aaron Courville, and Yoshua Bengio. 2014{\natexlab{a}}.
\newblock Generative adversarial nets.
\newblock In Z.~Ghahramani, M.~Welling, C.~Cortes, N.~D. Lawrence, and K.~Q.
  Weinberger, editors, \emph{Advances in Neural Information Processing Systems
  27}, pages 2672--2680. Curran Associates, Inc.

\bibitem[{Goodfellow et~al.(2014{\natexlab{b}})Goodfellow, Shlens, and
  Szegedy}]{goodfellow2014explaining}
Ian~J Goodfellow, Jonathon Shlens, and Christian Szegedy. 2014{\natexlab{b}}.
\newblock Explaining and harnessing adversarial examples.
\newblock \emph{arXiv preprint arXiv:1412.6572}.

\bibitem[{Gu et~al.(2018)Gu, Wang, Chen, Cho, and Li}]{gu2018meta}
Jiatao Gu, Yong Wang, Yun Chen, Kyunghyun Cho, and Victor~OK Li. 2018.
\newblock Meta-learning for low-resource neural machine translation.
\newblock \emph{arXiv preprint arXiv:1808.08437}.

\bibitem[{Higgins et~al.(2017)Higgins, Matthey, Pal, Burgess, Glorot,
  Botvinick, Mohamed, and Lerchner}]{Higgins2017betaVAELB}
I.~Higgins, Lo{\"i}c Matthey, A.~Pal, C.~Burgess, Xavier Glorot, M.~Botvinick,
  S.~Mohamed, and Alexander Lerchner. 2017.
\newblock beta-vae: Learning basic visual concepts with a constrained
  variational framework.
\newblock In \emph{ICLR}.

\bibitem[{Hjelm et~al.(2018)Hjelm, Fedorov, Lavoie-Marchildon, Grewal, Bachman,
  Trischler, and Bengio}]{hjelm2018mutualvae}
R~Devon Hjelm, Alex Fedorov, Samuel Lavoie-Marchildon, Karan Grewal, Phil
  Bachman, Adam Trischler, and Yoshua Bengio. 2018.
\newblock Learning deep representations by mutual information estimation and
  maximization.
\newblock \emph{arXiv preprint arXiv:1808.06670}.

\bibitem[{Hochreiter and Schmidhuber(1997)}]{hochreiter1997long}
Sepp Hochreiter and J{\"u}rgen Schmidhuber. 1997.
\newblock Long short-term memory.
\newblock \emph{Neural computation}, 9(8):1735--1780.

\bibitem[{Huang et~al.(2019)Huang, Stanforth, Welbl, Dyer, Yogatama, Gowal,
  Dvijotham, and Kohli}]{huang2019achieving}
Po-Sen Huang, Robert Stanforth, Johannes Welbl, Chris Dyer, Dani Yogatama, Sven
  Gowal, Krishnamurthy Dvijotham, and Pushmeet Kohli. 2019.
\newblock Achieving verified robustness to symbol substitutions via interval
  bound propagation.
\newblock \emph{arXiv preprint arXiv:1909.01492}.

\bibitem[{Iyyer et~al.(2018)Iyyer, Wieting, Gimpel, and
  Zettlemoyer}]{iyyer2018adversarial}
Mohit Iyyer, John Wieting, Kevin Gimpel, and Luke Zettlemoyer. 2018.
\newblock Adversarial example generation with syntactically controlled
  paraphrase networks.
\newblock \emph{arXiv preprint arXiv:1804.06059}.

\bibitem[{Jia et~al.(2019{\natexlab{a}})Jia, Liang, and Zhang}]{jia2019cross}
Chen Jia, Xiaobo Liang, and Yue Zhang. 2019{\natexlab{a}}.
\newblock Cross-domain ner using cross-domain language modeling.
\newblock In \emph{Proceedings of the 57th Annual Meeting of the Association
  for Computational Linguistics}, pages 2464--2474.

\bibitem[{Jia and Liang(2017)}]{jia2017adversarial}
Robin Jia and Percy Liang. 2017.
\newblock Adversarial examples for evaluating reading comprehension systems.
\newblock \emph{arXiv preprint arXiv:1707.07328}.

\bibitem[{Jia et~al.(2019{\natexlab{b}})Jia, Raghunathan, G{\"o}ksel, and
  Liang}]{jia2019certified}
Robin Jia, Aditi Raghunathan, Kerem G{\"o}ksel, and Percy Liang.
  2019{\natexlab{b}}.
\newblock Certified robustness to adversarial word substitutions.
\newblock \emph{arXiv preprint arXiv:1909.00986}.

\bibitem[{Joshi et~al.(2020)Joshi, Chen, Liu, Weld, Zettlemoyer, and
  Levy}]{danqi2020spanbert}
Mandar Joshi, Danqi Chen, Yinhan Liu, Daniel~S Weld, Luke Zettlemoyer, and Omer
  Levy. 2020.
\newblock Spanbert: Improving pre-training by representing and predicting
  spans.
\newblock \emph{Transactions of the Association for Computational Linguistics},
  8:64--77.

\bibitem[{Kalchbrenner et~al.(2014)Kalchbrenner, Grefenstette, and
  Blunsom}]{kalchbrenner2014convolutional}
Nal Kalchbrenner, Edward Grefenstette, and Phil Blunsom. 2014.
\newblock A convolutional neural network for modelling sentences.
\newblock \emph{arXiv preprint arXiv:1404.2188}.

\bibitem[{Kim and Mnih(2018)}]{kim2018factorvae}
Hyunjik Kim and Andriy Mnih. 2018.
\newblock Disentangling by factorising.
\newblock \emph{arXiv preprint arXiv:1802.05983}.

\bibitem[{Kim et~al.(2015)Kim, Stratos, Sarikaya, and
  Jeong}]{kim-etal-2015-new}
Young-Bum Kim, Karl Stratos, Ruhi Sarikaya, and Minwoo Jeong. 2015.
\newblock New transfer learning techniques for disparate label sets.
\newblock In \emph{Proceedings of the 53rd Annual Meeting of the Association
  for Computational Linguistics and the 7th International Joint Conference on
  Natural Language Processing (Volume 1: Long Papers)}, pages 473--482,
  Beijing, China. Association for Computational Linguistics.

\bibitem[{Kingma and Ba(2014)}]{kingma2014adam}
Diederik~P Kingma and Jimmy Ba. 2014.
\newblock Adam: A method for stochastic optimization.
\newblock \emph{arXiv preprint arXiv:1412.6980}.

\bibitem[{Kingma and Welling(2013)}]{kingma2013auto}
Diederik~P Kingma and Max Welling. 2013.
\newblock Auto-encoding variational bayes.
\newblock \emph{arXiv preprint arXiv:1312.6114}.

\bibitem[{Koh and Liang(2017)}]{koh2017understanding}
Pang~Wei Koh and Percy Liang. 2017.
\newblock Understanding black-box predictions via influence functions.
\newblock In \emph{Proceedings of the 34th International Conference on Machine
  Learning-Volume 70}, pages 1885--1894. JMLR. org.

\bibitem[{Kong et~al.(2019)Kong, d'Autume, Ling, Yu, Dai, and
  Yogatama}]{kong2019mutual}
Lingpeng Kong, Cyprien de~Masson d'Autume, Wang Ling, Lei Yu, Zihang Dai, and
  Dani Yogatama. 2019.
\newblock A mutual information maximization perspective of language
  representation learning.
\newblock \emph{arXiv preprint arXiv:1910.08350}.

\bibitem[{Krizhevsky et~al.(2012)Krizhevsky, Sutskever, and
  Hinton}]{krizhevsky2012imagenet}
Alex Krizhevsky, Ilya Sutskever, and Geoffrey~E Hinton. 2012.
\newblock Imagenet classification with deep convolutional neural networks.
\newblock In \emph{Advances in neural information processing systems}, pages
  1097--1105.

\bibitem[{Kumar et~al.(2018)Kumar, Sattigeri, and
  Balakrishnan}]{kumar2018variational}
Abhishek Kumar, Prasanna Sattigeri, and Avinash Balakrishnan. 2018.
\newblock Variational inference of disentangled latent concepts from unlabeled
  observations.
\newblock In \emph{International Conference on Learning Representations}.

\bibitem[{Lee et~al.(2018)Lee, Dernoncourt, and Szolovits}]{LEE18.878}
Ji~Young Lee, Franck Dernoncourt, and Peter Szolovits. 2018.
\newblock {Transfer Learning for Named-Entity Recognition with Neural
  Networks}.
\newblock In \emph{Proceedings of the Eleventh International Conference on
  Language Resources and Evaluation (LREC 2018)}, Miyazaki, Japan. European
  Language Resources Association (ELRA).

\bibitem[{Lei et~al.(2016)Lei, Barzilay, and Jaakkola}]{lei2016rationalizing}
Tao Lei, Regina Barzilay, and Tommi Jaakkola. 2016.
\newblock Rationalizing neural predictions.
\newblock \emph{arXiv preprint arXiv:1606.04155}.

\bibitem[{Levy et~al.(2017)Levy, Seo, Choi, and Zettlemoyer}]{levy2017zero}
Omer Levy, Minjoon Seo, Eunsol Choi, and Luke Zettlemoyer. 2017.
\newblock Zero-shot relation extraction via reading comprehension.
\newblock \emph{arXiv preprint arXiv:1706.04115}.

\bibitem[{Li et~al.(2015)Li, Chen, Hovy, and Jurafsky}]{li2015visualizing}
Jiwei Li, Xinlei Chen, Eduard Hovy, and Dan Jurafsky. 2015.
\newblock Visualizing and understanding neural models in nlp.
\newblock \emph{arXiv preprint arXiv:1506.01066}.

\bibitem[{Li et~al.(2016)Li, Monroe, and Jurafsky}]{li2016understanding}
Jiwei Li, Will Monroe, and Dan Jurafsky. 2016.
\newblock Understanding neural networks through representation erasure.
\newblock \emph{arXiv preprint arXiv:1612.08220}.

\bibitem[{Li et~al.(2014)Li, Ott, Cardie, and Hovy}]{li2014towards}
Jiwei Li, Myle Ott, Claire Cardie, and Eduard Hovy. 2014.
\newblock Towards a general rule for identifying deceptive opinion spam.
\newblock In \emph{Proceedings of the 52nd Annual Meeting of the Association
  for Computational Linguistics (Volume 1: Long Papers)}, pages 1566--1576.

\bibitem[{Li et~al.(2018)Li, Jia, He, and Liang}]{li2018delete}
Juncen Li, Robin Jia, He~He, and Percy Liang. 2018.
\newblock Delete, retrieve, generate: A simple approach to sentiment and style
  transfer.
\newblock \emph{arXiv preprint arXiv:1804.06437}.

\bibitem[{Li and Eisner(2019)}]{li2019specializing}
Xiang~Lisa Li and Jason Eisner. 2019.
\newblock Specializing word embeddings (for parsing) by information bottleneck.
\newblock \emph{arXiv preprint arXiv:1910.00163}.

\bibitem[{Li et~al.(2019{\natexlab{a}})Li, Feng, Meng, Han, Wu, and
  Li}]{li2019unified}
Xiaoya Li, Jingrong Feng, Yuxian Meng, Qinghong Han, Fei Wu, and Jiwei Li.
  2019{\natexlab{a}}.
\newblock A unified mrc framework for named entity recognition.
\newblock \emph{arXiv preprint arXiv:1910.11476}.

\bibitem[{Li et~al.(2019{\natexlab{b}})Li, Li, Wei, Bing, Zhang, and
  Yang}]{li-etal-2019-transferable}
Zheng Li, Xin Li, Ying Wei, Lidong Bing, Yu~Zhang, and Qiang Yang.
  2019{\natexlab{b}}.
\newblock Transferable end-to-end aspect-based sentiment analysis with
  selective adversarial learning.
\newblock In \emph{Proceedings of the 2019 Conference on Empirical Methods in
  Natural Language Processing and the 9th International Joint Conference on
  Natural Language Processing (EMNLP-IJCNLP)}, pages 4590--4600, Hong Kong,
  China. Association for Computational Linguistics.

\bibitem[{Liang et~al.(2017)Liang, Li, Su, Bian, Li, and Shi}]{liang2017deep}
Bin Liang, Hongcheng Li, Miaoqiang Su, Pan Bian, Xirong Li, and Wenchang Shi.
  2017.
\newblock Deep text classification can be fooled.
\newblock \emph{arXiv preprint arXiv:1704.08006}.

\bibitem[{Lin and Lu(2018)}]{lin-lu-2018-neural}
Bill~Yuchen Lin and Wei Lu. 2018.
\newblock Neural adaptation layers for cross-domain named entity recognition.
\newblock In \emph{Proceedings of the 2018 Conference on Empirical Methods in
  Natural Language Processing}, pages 2012--2022, Brussels, Belgium.
  Association for Computational Linguistics.

\bibitem[{Linzen et~al.(2016)Linzen, Dupoux, and
  Goldberg}]{linzen2016assessing}
Tal Linzen, Emmanuel Dupoux, and Yoav Goldberg. 2016.
\newblock Assessing the ability of lstms to learn syntax-sensitive
  dependencies.
\newblock \emph{Transactions of the Association for Computational Linguistics},
  4:521--535.

\bibitem[{Liu et~al.(2019)Liu, Ott, Goyal, Du, Joshi, Chen, Levy, Lewis,
  Zettlemoyer, and Stoyanov}]{yinhan2019roberta}
Yinhan Liu, Myle Ott, Naman Goyal, Jingfei Du, Mandar Joshi, Danqi Chen, Omer
  Levy, Mike Lewis, Luke Zettlemoyer, and Veselin Stoyanov. 2019.
\newblock Roberta: A robustly optimized bert pretraining approach.
\newblock \emph{arXiv preprint arXiv:1907.11692}.

\bibitem[{Locatello et~al.(2019)Locatello, Bauer, Lucic, Raetsch, Gelly,
  Sch{\"o}lkopf, and Bachem}]{pmlr-v97-locatello19a}
Francesco Locatello, Stefan Bauer, Mario Lucic, Gunnar Raetsch, Sylvain Gelly,
  Bernhard Sch{\"o}lkopf, and Olivier Bachem. 2019.
\newblock Challenging common assumptions in the unsupervised learning of
  disentangled representations.
\newblock In \emph{Proceedings of the 36th International Conference on Machine
  Learning}, volume~97 of \emph{Proceedings of Machine Learning Research},
  pages 4114--4124, Long Beach, California, USA. PMLR.

\bibitem[{Luong and Manning(2015)}]{Luong-Manning:iwslt15}
Minh-Thang Luong and Christopher~D. Manning. 2015.
\newblock Stanford neural machine translation systems for spoken language
  domain.
\newblock In \emph{International Workshop on Spoken Language Translation}, Da
  Nang, Vietnam.

\bibitem[{Meng et~al.(2019{\natexlab{a}})Meng, Ren, Sun, Li, Yuan, Wu, and
  Li}]{yuxian2019largenmt}
Yuxian Meng, Xiangyuan Ren, Zijun Sun, Xiaoya Li, Arianna Yuan, Fei Wu, and
  Jiwei Li. 2019{\natexlab{a}}.
\newblock Large-scale pretraining for neural machine translation with tens of
  billions of sentence pairs.
\newblock \emph{arXiv preprint arXiv:1909.11861}.

\bibitem[{Meng et~al.(2019{\natexlab{b}})Meng, Wu, Wang, Li, Nie, Yin, Li, Han,
  Sun, and Li}]{yuxian2019glyce}
Yuxian Meng, Wei Wu, Fei Wang, Xiaoya Li, Ping Nie, Fan Yin, Muyu Li, Qinghong
  Han, Xiaofei Sun, and Jiwei Li. 2019{\natexlab{b}}.
\newblock Glyce: Glyph-vectors for chinese character representations.
\newblock In \emph{Advances in Neural Information Processing Systems}, pages
  2742--2753.

\bibitem[{Mikolov et~al.(2010)Mikolov, Karafi{\'a}t, Burget,
  {\v{C}}ernock{\`y}, and Khudanpur}]{mikolov2010recurrent}
Tom{\'a}{\v{s}} Mikolov, Martin Karafi{\'a}t, Luk{\'a}{\v{s}} Burget, Jan
  {\v{C}}ernock{\`y}, and Sanjeev Khudanpur. 2010.
\newblock Recurrent neural network based language model.
\newblock In \emph{Eleventh annual conference of the international speech
  communication association}.

\bibitem[{Mirza and Osindero(2014)}]{mirza2014conditional}
Mehdi Mirza and Simon Osindero. 2014.
\newblock Conditional generative adversarial nets.
\newblock \emph{arXiv preprint arXiv:1411.1784}.

\bibitem[{Miyato et~al.(2016)Miyato, Dai, and
  Goodfellow}]{miyato2016adversarial}
Takeru Miyato, Andrew~M. Dai, and Ian Goodfellow. 2016.
\newblock Adversarial training methods for semi-supervised text classification.

\bibitem[{Mou et~al.(2016)Mou, Meng, Yan, Li, Xu, Zhang, and
  Jin}]{mou2016transferable}
Lili Mou, Zhao Meng, Rui Yan, Ge~Li, Yan Xu, Lu~Zhang, and Zhi Jin. 2016.
\newblock How transferable are neural networks in nlp applications?
\newblock \emph{arXiv preprint arXiv:1603.06111}.

\bibitem[{Nguyen et~al.(2015)Nguyen, Yosinski, and Clune}]{nguyen2015deep}
Anh Nguyen, Jason Yosinski, and Jeff Clune. 2015.
\newblock Deep neural networks are easily fooled: High confidence predictions
  for unrecognizable images.
\newblock In \emph{Proceedings of the IEEE conference on computer vision and
  pattern recognition}, pages 427--436.

\bibitem[{{Papernot} et~al.(2016{\natexlab{a}}){Papernot}, {McDaniel}, {Jha},
  {Fredrikson}, {Celik}, and {Swami}}]{7467366}
N.~{Papernot}, P.~{McDaniel}, S.~{Jha}, M.~{Fredrikson}, Z.~B. {Celik}, and
  A.~{Swami}. 2016{\natexlab{a}}.
\newblock The limitations of deep learning in adversarial settings.
\newblock In \emph{2016 IEEE European Symposium on Security and Privacy (EuroS
  P)}, pages 372--387.

\bibitem[{{Papernot} et~al.(2016{\natexlab{b}}){Papernot}, {McDaniel}, {Swami},
  and {Harang}}]{7795300}
N.~{Papernot}, P.~{McDaniel}, A.~{Swami}, and R.~{Harang}. 2016{\natexlab{b}}.
\newblock Crafting adversarial input sequences for recurrent neural networks.
\newblock In \emph{MILCOM 2016 - 2016 IEEE Military Communications Conference},
  pages 49--54.

\bibitem[{Papernot et~al.(2017)Papernot, McDaniel, Goodfellow, Jha, Celik, and
  Swami}]{papernot2017practical}
Nicolas Papernot, Patrick McDaniel, Ian Goodfellow, Somesh Jha, Z~Berkay Celik,
  and Ananthram Swami. 2017.
\newblock Practical black-box attacks against machine learning.
\newblock In \emph{Proceedings of the 2017 ACM on Asia conference on computer
  and communications security}, pages 506--519.

\bibitem[{Papernot et~al.(2016)Papernot, McDaniel, Swami, and
  Harang}]{papernot2016crafting}
Nicolas Papernot, Patrick McDaniel, Ananthram Swami, and Richard Harang. 2016.
\newblock Crafting adversarial input sequences for recurrent neural networks.
\newblock In \emph{MILCOM 2016-2016 IEEE Military Communications Conference},
  pages 49--54. IEEE.

\bibitem[{Papineni et~al.(2002)Papineni, Roukos, Ward, and
  Zhu}]{papineni-etal-2002-bleu}
Kishore Papineni, Salim Roukos, Todd Ward, and Wei-Jing Zhu. 2002.
\newblock {B}leu: a method for automatic evaluation of machine translation.
\newblock In \emph{Proceedings of the 40th Annual Meeting of the Association
  for Computational Linguistics}, pages 311--318, Philadelphia, Pennsylvania,
  USA. Association for Computational Linguistics.

\bibitem[{Patel et~al.(2014)Patel, Gopalan, Li, and
  Chellappa}]{patel2014visual}
Vishal~M Patel, Raghuraman Gopalan, Ruonan Li, and Rama Chellappa. 2014.
\newblock Visual domain adaptation: An overview of recent advances.
\newblock \emph{IEEE Signal Processing Magazine}, 2.

\bibitem[{Ren et~al.(2019{\natexlab{a}})Ren, Deng, He, and
  Che}]{ren-etal-2019-generating}
Shuhuai Ren, Yihe Deng, Kun He, and Wanxiang Che. 2019{\natexlab{a}}.
\newblock Generating natural language adversarial examples through probability
  weighted word saliency.
\newblock In \emph{Proceedings of the 57th Annual Meeting of the Association
  for Computational Linguistics}, pages 1085--1097, Florence, Italy.
  Association for Computational Linguistics.

\bibitem[{Ren et~al.(2019{\natexlab{b}})Ren, Deng, He, and
  Che}]{ren2019generating}
Shuhuai Ren, Yihe Deng, Kun He, and Wanxiang Che. 2019{\natexlab{b}}.
\newblock Generating natural language adversarial examples through probability
  weighted word saliency.
\newblock In \emph{Proceedings of the 57th annual meeting of the association
  for computational linguistics}, pages 1085--1097.

\bibitem[{Ribeiro et~al.(2018)Ribeiro, Singh, and
  Guestrin}]{ribeiro2018semantically}
Marco~Tulio Ribeiro, Sameer Singh, and Carlos Guestrin. 2018.
\newblock Semantically equivalent adversarial rules for debugging nlp models.
\newblock In \emph{Proceedings of the 56th Annual Meeting of the Association
  for Computational Linguistics (Volume 1: Long Papers)}, pages 856--865.

\bibitem[{Ruder(2019)}]{Ruder2019Neural}
Sebastian Ruder. 2019.
\newblock \emph{Neural Transfer Learning for Natural Language Processing}.
\newblock Ph.D. thesis, National University of Ireland, Galway.

\bibitem[{Sato et~al.(2018)Sato, Suzuki, Shindo, and Matsumoto}]{ijcai2018-601}
Motoki Sato, Jun Suzuki, Hiroyuki Shindo, and Yuji Matsumoto. 2018.
\newblock Interpretable adversarial perturbation in input embedding space for
  text.
\newblock In \emph{Proceedings of the Twenty-Seventh International Joint
  Conference on Artificial Intelligence, {IJCAI-18}}, pages 4323--4330.
  International Joint Conferences on Artificial Intelligence Organization.

\bibitem[{Sennrich et~al.(2016{\natexlab{a}})Sennrich, Haddow, and
  Birch}]{sennrich2016back-translation}
Rico Sennrich, Barry Haddow, and Alexandra Birch. 2016{\natexlab{a}}.
\newblock Improving neural machine translation models with monolingual data.
\newblock In \emph{Proceedings of the 54th Annual Meeting of the Association
  for Computational Linguistics (Volume 1: Long Papers)}, pages 86--96, Berlin,
  Germany. Association for Computational Linguistics.

\bibitem[{Sennrich et~al.(2016{\natexlab{b}})Sennrich, Haddow, and
  Birch}]{sennrich-etal-2016-neural}
Rico Sennrich, Barry Haddow, and Alexandra Birch. 2016{\natexlab{b}}.
\newblock Neural machine translation of rare words with subword units.
\newblock In \emph{Proceedings of the 54th Annual Meeting of the Association
  for Computational Linguistics (Volume 1: Long Papers)}, pages 1715--1725,
  Berlin, Germany. Association for Computational Linguistics.

\bibitem[{Seo et~al.(2016)Seo, Kembhavi, Farhadi, and Hajishirzi}]{bidaf}
Min~Joon Seo, Aniruddha Kembhavi, Ali Farhadi, and Hannaneh Hajishirzi. 2016.
\newblock Bidirectional attention flow for machine comprehension.
\newblock \emph{CoRR}, abs/1611.01603.

\bibitem[{Shi et~al.(2020)Shi, Zhang, Chang, Huang, and
  Hsieh}]{Shi2020Robustness}
Zhouxing Shi, Huan Zhang, Kai-Wei Chang, Minlie Huang, and Cho-Jui Hsieh. 2020.
\newblock Robustness verification for transformers.
\newblock In \emph{International Conference on Learning Representations}.

\bibitem[{Shwartz-Ziv and Tishby(2017)}]{shwartz2017opening}
Ravid Shwartz-Ziv and Naftali Tishby. 2017.
\newblock Opening the black box of deep neural networks via information.
\newblock \emph{arXiv preprint arXiv:1703.00810}.

\bibitem[{Simonyan et~al.(2013)Simonyan, Vedaldi, and
  Zisserman}]{simonyan2013deep}
Karen Simonyan, Andrea Vedaldi, and Andrew Zisserman. 2013.
\newblock Deep inside convolutional networks: Visualising image classification
  models and saliency maps.
\newblock \emph{arXiv preprint arXiv:1312.6034}.

\bibitem[{Steeg(2017)}]{steeg2017unsupervised}
Greg~Ver Steeg. 2017.
\newblock Unsupervised learning via total correlation explanation.
\newblock \emph{arXiv preprint arXiv:1706.08984}.

\bibitem[{Sun et~al.(2016)Sun, Feng, and Saenko}]{AAAI1612443}
Baochen Sun, Jiashi Feng, and Kate Saenko. 2016.
\newblock Return of frustratingly easy domain adaptation.

\bibitem[{Sutskever et~al.(2014)Sutskever, Vinyals, and
  Le}]{sutskever2014sequence}
Ilya Sutskever, Oriol Vinyals, and Quoc~V Le. 2014.
\newblock Sequence to sequence learning with neural networks.
\newblock In \emph{Advances in neural information processing systems}, pages
  3104--3112.

\bibitem[{Szegedy et~al.(2013)Szegedy, Zaremba, Sutskever, Bruna, Erhan,
  Goodfellow, and Fergus}]{szegedy2013intriguing}
Christian Szegedy, Wojciech Zaremba, Ilya Sutskever, Joan Bruna, Dumitru Erhan,
  Ian Goodfellow, and Rob Fergus. 2013.
\newblock Intriguing properties of neural networks.
\newblock \emph{arXiv preprint arXiv:1312.6199}.

\bibitem[{Tan et~al.(2009)Tan, Cheng, Wang, and Xu}]{tan2009adapting}
Songbo Tan, Xueqi Cheng, Yuefen Wang, and Hongbo Xu. 2009.
\newblock Adapting naive bayes to domain adaptation for sentiment analysis.
\newblock In \emph{European Conference on Information Retrieval}, pages
  337--349. Springer.

\bibitem[{Tishby et~al.(2000)Tishby, Pereira, and Bialek}]{info2000bottleneck}
Naftali Tishby, Fernando C.~N. Pereira, and William Bialek. 2000.
\newblock The information bottleneck method.
\newblock \emph{CoRR}, physics/0004057.

\bibitem[{Tishby and Zaslavsky(2015)}]{tishby2015dl}
Naftali Tishby and Noga Zaslavsky. 2015.
\newblock Deep learning and the information bottleneck principle.
\newblock In \emph{2015 IEEE Information Theory Workshop (ITW)}, pages 1--5.
  IEEE.

\bibitem[{Ul~Haq et~al.(2020)Ul~Haq, Abdul~Rauf, Shoukat, and
  Hira}]{ul-haq-etal-2020-improving}
Sami Ul~Haq, Sadaf Abdul~Rauf, Arslan Shoukat, and Noor-e Hira. 2020.
\newblock Improving document-level neural machine translation with domain
  adaptation.
\newblock In \emph{Proceedings of the Fourth Workshop on Neural Generation and
  Translation}, pages 225--231, Online. Association for Computational
  Linguistics.

\bibitem[{Vaswani et~al.(2017{\natexlab{a}})Vaswani, Shazeer, Parmar,
  Uszkoreit, Jones, Gomez, Kaiser, and Polosukhin}]{vaswani2017transformer}
Ashish Vaswani, Noam Shazeer, Niki Parmar, Jakob Uszkoreit, Llion Jones,
  Aidan~N Gomez, \L~ukasz Kaiser, and Illia Polosukhin. 2017{\natexlab{a}}.
\newblock Attention is all you need.
\newblock In I.~Guyon, U.~V. Luxburg, S.~Bengio, H.~Wallach, R.~Fergus,
  S.~Vishwanathan, and R.~Garnett, editors, \emph{Advances in Neural
  Information Processing Systems 30}, pages 5998--6008. Curran Associates, Inc.

\bibitem[{Vaswani et~al.(2017{\natexlab{b}})Vaswani, Shazeer, Parmar,
  Uszkoreit, Jones, Gomez, Kaiser, and Polosukhin}]{attentionisallyouneed}
Ashish Vaswani, Noam Shazeer, Niki Parmar, Jakob Uszkoreit, Llion Jones,
  Aidan~N. Gomez, Lukasz Kaiser, and Illia Polosukhin. 2017{\natexlab{b}}.
\newblock Attention is all you need.
\newblock \emph{CoRR}, abs/1706.03762.

\bibitem[{Ver~Steeg and Galstyan(2015)}]{ver2015maximally}
Greg Ver~Steeg and Aram Galstyan. 2015.
\newblock Maximally informative hierarchical representations of
  high-dimensional data.
\newblock In \emph{Artificial Intelligence and Statistics}, pages 1004--1012.

\bibitem[{Xie et~al.(2017)Xie, Wang, Li, L{\'e}vy, Nie, Jurafsky, and
  Ng}]{xie2017data}
Ziang Xie, Sida~I Wang, Jiwei Li, Daniel L{\'e}vy, Aiming Nie, Dan Jurafsky,
  and Andrew~Y Ng. 2017.
\newblock Data noising as smoothing in neural network language models.
\newblock \emph{arXiv preprint arXiv:1703.02573}.

\bibitem[{Yang et~al.(2018)Yang, Liang, and Zhang}]{yang-etal-2018-design}
Jie Yang, Shuailong Liang, and Yue Zhang. 2018.
\newblock Design challenges and misconceptions in neural sequence labeling.
\newblock In \emph{Proceedings of the 27th International Conference on
  Computational Linguistics}, pages 3879--3889, Santa Fe, New Mexico, USA.
  Association for Computational Linguistics.

\bibitem[{Yosinski et~al.(2014)Yosinski, Clune, Bengio, and
  Lipson}]{yosinski2014transferable}
Jason Yosinski, Jeff Clune, Yoshua Bengio, and Hod Lipson. 2014.
\newblock How transferable are features in deep neural networks?
\newblock In \emph{Advances in neural information processing systems}, pages
  3320--3328.

\bibitem[{Yu et~al.(2018)Yu, Dohan, Luong, Zhao, Chen, Norouzi, and
  Le}]{yu2018qanet}
Adams~Wei Yu, David Dohan, Minh-Thang Luong, Rui Zhao, Kai Chen, Mohammad
  Norouzi, and Quoc~V Le. 2018.
\newblock Qanet: Combining local convolution with global self-attention for
  reading comprehension.
\newblock \emph{arXiv preprint arXiv:1804.09541}.

\bibitem[{Yuan et~al.(2019)Yuan, He, Zhu, and Li}]{yuan2019adversarial}
Xiaoyong Yuan, Pan He, Qile Zhu, and Xiaolin Li. 2019.
\newblock Adversarial examples: Attacks and defenses for deep learning.
\newblock \emph{IEEE transactions on neural networks and learning systems},
  30(9):2805--2824.

\bibitem[{Zeiler and Fergus(2014)}]{zeiler2014visualizing}
Matthew~D Zeiler and Rob Fergus. 2014.
\newblock Visualizing and understanding convolutional networks.
\newblock In \emph{European conference on computer vision}, pages 818--833.
  Springer.

\bibitem[{Zhao et~al.(2017)Zhao, Dua, and Singh}]{zhao2017generating}
Zhengli Zhao, Dheeru Dua, and Sameer Singh. 2017.
\newblock Generating natural adversarial examples.
\newblock \emph{arXiv preprint arXiv:1710.11342}.

\bibitem[{Zhou et~al.(2020)Zhou, Zheng, Hsieh, Chang, and
  Huang}]{zhou2020defense}
Yi~Zhou, Xiaoqing Zheng, Cho-Jui Hsieh, Kai-wei Chang, and Xuanjing Huang.
  2020.
\newblock Defense against adversarial attacks in nlp via dirichlet neighborhood
  ensemble.
\newblock \emph{arXiv preprint arXiv:2006.11627}.

\bibitem[{Zhu et~al.(2020)Zhu, Cheng, Gan, Sun, Goldstein, and
  Liu}]{Zhu2020FreeLB}
Chen Zhu, Yu~Cheng, Zhe Gan, Siqi Sun, Tom Goldstein, and Jingjing Liu. 2020.
\newblock Freelb: Enhanced adversarial training for natural language
  understanding.
\newblock In \emph{International Conference on Learning Representations}.

\end{thebibliography}
\bibliographystyle{acl_natbib}

\newpage
\appendix
\section{Derivation of Variational Information Bottleneck}
\label{appendix1}
Below we take the derivation of VIB from \citet{alemi2016vib}.

We first decompose the joint distribution $p(X,Y,Z)$ into:
\begin{equation}
  \begin{aligned}
    p(X,Y,Z)&=p(X)p(Y|X)p(Z|X,Y)\\
    &=p(X)p(Y|X)p(Z|X)
  \end{aligned}
\end{equation}
Then, for the first term in the IB objective $I(Z,Y)-\beta I(Z,X)$, we write it out in full:
\begin{equation}
  \begin{aligned}
    I(Z,Y)&=\int p(y,z)\log\frac{p(y,z)}{p(y)p(z)}\ \mathrm{d}y\mathrm{d}z\\
    &=\int  p(y,z)\log\frac{p(y|z)}{p(y)}\ \mathrm{d}y\mathrm{d}z
  \end{aligned}
\end{equation}
where $p(y|z)$ is fully defined by the encoder and the Markov Chain as follows:
\begin{equation}
  \begin{aligned}
    p(y|z)&=\int p(x,y|z)\ \mathrm{d}x\\
    &=\int p(y|x)p(y|x)\ \mathrm{d}x\\
    &=\int\frac{p(y|x)p(z|x)p(x)}{p(z)}\ \mathrm{d}x
  \end{aligned}
\end{equation}
Let $q(y|z)$ be a variational approximation to $p(y|z)$. By the fact that the KL divergence is non-negative, we have:
\begin{equation}
  \begin{aligned}
    &D_\text{KL}(p(Y|Z),q(Y|Z))\ge 0 \Longrightarrow \\
    &\int p(y|z)\log p(y|z)\ \mathrm{d}y\ge \int p(y|z)\log q(y|z)\ \mathrm{d}y
  \end{aligned}
\end{equation}
and hence 
\begin{equation}
  \begin{aligned}
    &I(Z,Y)\ge \int p(y,z)\log\frac{q(y|z)}{p(y)}\  \mathrm{d}y \mathrm{d}z\\
    &=\int  p(y,z)\log q(y,z)\ \mathrm{d}y \mathrm{d}z-\int  p(y)\log p(y)\ \mathrm{d}y\\
    &=\int  p(y,z)\log q(y,z)\ \mathrm{d}y \mathrm{d}z+H(Y)
  \end{aligned}
\end{equation}
We omit the second term and rewrite $p(y,z)$ as:
\begin{equation}
\begin{aligned}
p(y,z)&=\int  p(x,y,z)\ \mathrm{d}x\\
&=\int  p(x)p(y|x)p(z|x)\ \mathrm{d}x
\end{aligned}
\end{equation}
which gives:
\begin{equation}
  \begin{aligned}
    I(Z,Y)\ge \int  p(x)p(y|x)p(z|x)\log q(y|z)\ \mathrm{d}x\mathrm{d}y\mathrm{d}z
  \end{aligned}
\end{equation}

For the term $\beta I(Z,X)$, we can Similarly expand it as:
\begin{equation}
  \begin{aligned}
    I(Z,X)&=\int  p(x,z)\log \frac{p(z|x)}{p(z)}\ \mathrm{d}z\mathrm{d}x\\
    &=\int  p(x,z)\log p(z|x)\ \mathrm{d}z\mathrm{d}x\\
    &-\int  p(z)\log p(z)\ \mathrm{d}z
  \end{aligned}
\end{equation}
 Computing $p(z)$ is intractable, so we introduce a variational approximation $r(z)$ to it. Again using the fact that the KL divergence is non-negative, we have:
 \begin{equation}
  \begin{aligned}
    I(Z,X)\le\int p(x)p(z|x)\log \frac{p(z|x)}{r(z)}\  \mathrm{d}x\mathrm{d}z 
  \end{aligned}
\end{equation}
At last we have  that:
\begin{equation}
  \begin{aligned}
    &I(Z,Y)-\beta I(Z,X)\\
    &\ge \int p(x)p(y|x)p(z|x)\log q(y|z)\ \mathrm{d}x\mathrm{d}y\mathrm{d}z\\
    &-\beta \int  p(x)p(z|x)\log \frac{p(z|x)}{r(z)}\ \mathrm{d}x\mathrm{d}z\\
    &\triangleq\mathcal{L}_\text{VIB}
  \end{aligned}
\end{equation}
To compute $p(x,y)$ we can use the empirical data distribution $p(x,y)=\frac{1}{N}\sum_{n=1}^N\delta_{x_n}(x)\delta_{y_n}(y)$, and hence we can derive the final formula with the reparameterization trick $p(z|x)\mathrm{d}z=p(\epsilon)\mathrm{d}\epsilon$: 
\begin{equation}
  \begin{aligned}
    \mathcal{L}_\text{VIB}\triangleq&\frac{1}{N}\sum_{n=1}^N\mathbb{E}_{p(\mathbf{\epsilon})}[-\log q(y_n|f(x_n,\mathbf{\epsilon}))]\\
    &+\beta D_\text{KL}(p(z|x_n),r(z))
  \end{aligned}
  \end{equation}
which is exactly Eq.\ref{eq3}.
\end{document}